\documentclass[10pt,twocolumn,letterpaper]{article}

\usepackage{cvpr}              
\definecolor{cvprblue}{rgb}{0.21,0.49,0.74}
\usepackage[pagebackref,breaklinks,colorlinks,allcolors=cvprblue]{hyperref}

\def\confName{CVPR}
\def\confYear{2027}

\newcommand{\benchname}{{ENC-QA}\xspace}
\newcommand{\benchsize}{{77,312}\xspace}
\newcommand{\noaasize}{{6,612}\xspace}

\newcommand{\progressivesize}{{8,227}\xspace}

\usepackage{xcolor}
\usepackage{textcomp}
\usepackage{multirow}
\usepackage[most]{tcolorbox}
\usepackage{fvextra}
\usepackage{xcolor}

\usepackage[T1]{fontenc}
\usepackage{textcomp}

\definecolor{promptbg}{RGB}{247,247,247}
\definecolor{promptborder}{RGB}{205,205,205}\usepackage[most]{tcolorbox}
\usepackage{fvextra}
\usepackage{xcolor}
\usepackage{cuted}
\usepackage[most]{tcolorbox}

\definecolor{promptbg}{RGB}{247,247,247}
\definecolor{promptborder}{RGB}{205,205,205}

\newtcblisting{judgeprompt}[1][]{
    enhanced,
    breakable,
    colback=promptbg,
    colframe=promptborder,
    boxrule=0.4pt,
    arc=2pt,
    left=8pt,
    right=8pt,
    top=8pt,
    bottom=8pt,
    listing only,
    listing engine=listings,
    listing options={
        basicstyle=\ttfamily\scriptsize,
        breaklines=true,
        breakatwhitespace=true,
        columns=fullflexible,
        keepspaces=true,
        showstringspaces=false,
        tabsize=2
    },
    #1
}

\title{\benchname: A Diagnostic Benchmark for Vision-Language Models on Electronic Navigational Charts}

\author{Jiyoon Pyo,\textsuperscript{1}\quad
        Jacob Arndt,\textsuperscript{2}\quad
        Yao-Yi Chiang,\textsuperscript{1}\quad
        Dalton Lunga\textsuperscript{2}\quad\\ \\
        \textsuperscript{1}University of Minnesota - Twin Cities
        \qquad
        \textsuperscript{2}Oak Ridge National Laboratory\\
        {\tt\small
        \{py000005,yaoyi\}@umn.edu \qquad
        \{arndtjw,lungadd\}@ornl.gov
        }
}

\begin{document}
\maketitle
\begin{abstract}

Nautical charts encode safety-critical maritime information through standardized symbols, attributes, text, and heterogeneous geometries. Interpreting these charts and assessing the navigational significance of successive chart updates requires recognizing nautical features, determining their spatial relationships, and applying navigational knowledge. Vision-language models (VLMs) offer a promising approach to nautical-chart interpretation, yet systematic evaluation of their fine-grained nautical feature recognition, chart-based spatial reasoning, and application of navigational knowledge to assessing chart changes remains limited. We introduce \benchname, a diagnostic benchmark comprising \benchsize questions derived from \noaasize National Oceanic and Atmospheric Administration (NOAA) Electronic Navigational Charts (ENCs). The benchmark evaluates four complementary capabilities: ENC Perception, Spatial Reasoning, Joint Perception and Spatial Reasoning, and Application. Of these, \progressivesize questions form a progressive subset that decomposes selected chart-change cases into linked questions progressing from change detection to criticality assessment, enabling case-level analysis of how models use perceptual and spatial evidence to determine the navigational significance of chart changes. Evaluations of 10 open-weight VLMs reveal substantial limitations in fine-grained nautical feature recognition and feature-grounded spatial reasoning. Controlled evidence interventions show that relevant perceptual and spatial information improves change-criticality accuracy. Our findings establish \benchname as a diagnostic resource for nautical-chart understanding and safety-critical navigational decisions. 

\end{abstract}    
\begin{figure*}[t]
    \centering
    \includegraphics[width=\linewidth]{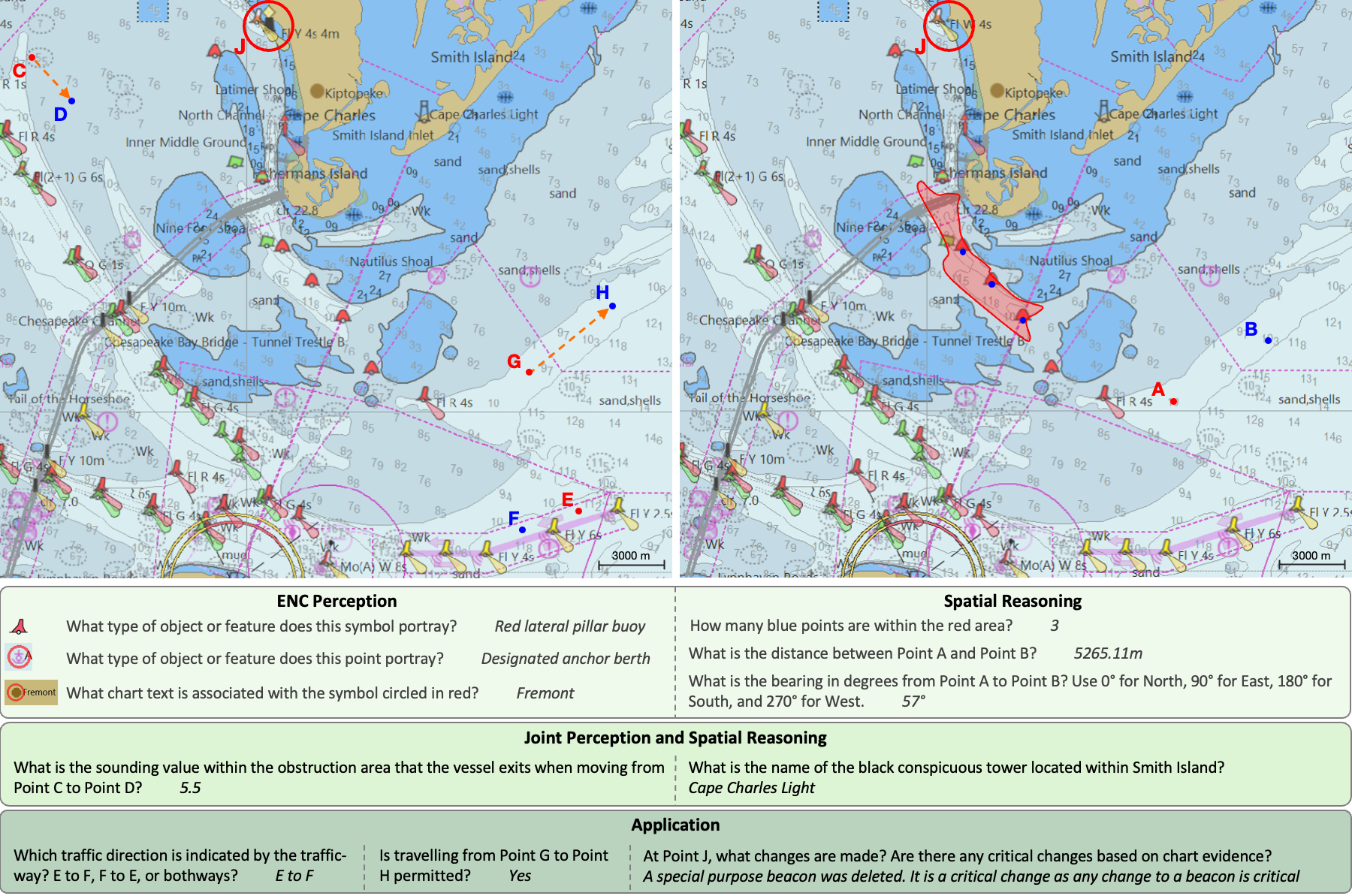}
    \caption{\textbf{Overview of \benchname}. The benchmark evaluates the capabilities required for nautical chart automation across four complementary capability categories: ENC Perception, Spatial Reasoning, Joint Perception and Spatial Reasoning, and Application. 
    }
    \label{fig:teaser}
    \vspace{-6pt}
\end{figure*}

\section{Introduction}
Nautical charts provide information essential for safe maritime navigation, including water depths, navigational hazards, regulated areas, traffic routes, and aids to navigation~\citep{noaa_nautical_cartography, iho_s57_enc_production}. They convey this information using internationally standardized symbols and portrayal conventions~\citep{iho-s52, iho-s57-appb}. Interpreting these symbols requires knowledge of the nautical features they represent, as visually similar symbols may denote features with different operational roles or navigational restrictions. The navigational significance of a nautical feature also depends on spatial relationships, such as proximity to routes or intersections with restricted areas~\citep{iho_s4_2021, imo-msc232-2006, imo_ecdis}. Accurate nautical-chart interpretation requires integrating domain-specific semantic knowledge with spatial reasoning to assess the navigational implications of features represented on the chart.

Nautical charts are regularly updated to reflect changes in hydrographic conditions and navigational information~\citep{noaa-chart-update}. Assessing the navigational significance of these updates requires both recognizing changes in nautical features and considering their spatial relationships with surrounding features. For example, a change in a buoy's light \textit{color} may affect how navigators determine which side of the buoy to pass, even if its location remains unchanged~\citep{iho_s4_2021}. Similarly, a new obstruction or a reduction in water depth \textit{within} a navigation route may introduce safety risks~\citep{imo_ecdis, imo-msc232-2006}. Timely determination of the criticality of such changes requires identifying nautical features, understanding their spatial relationships, and applying domain-specific safety criteria, motivating a benchmarking protocol to evaluate these capabilities individually and jointly. 

Modern maritime navigation increasingly distributes nautical chart information as Electronic Navigational Charts (ENCs), which encode navigational features and their attributes in a structured vector format~\citep{noaa_enc, iho-s57-appb}. Through standardized portrayal rules~\citep{iho-s52, iho_s57_uoc_2024}, Electronic Chart Display and Information Systems (ECDIS)~\citep{imo_ecdis} transform these vector-encoded features into a visual chart representation consisting of nautical symbols, labels, and other graphical elements, allowing navigators to interpret the structured information in chart form. However, navigational information is not always available as ENCs. In some offshore and international waters, limited ENC coverage still requires reliance on raster navigational charts~\citep{imo_ecdis_guidance_2022, iho_s66, iho_s61}, and many historical nautical charts exist only as scanned raster images~\citep{noaa_historical_map_chart, noaa_historical_chart_collection}. Moreover, time-sensitive navigational updates may initially be communicated through visual diagrams in notices to navigators before being incorporated into updated vector charts~\citep{iho_s4_2021}. These settings highlight the importance of recovering and reasoning over navigational information directly from rasterized charts, rather than relying exclusively on access to structured vector data.

Existing approaches~\citep{arndt-2026-enc} assess ENC change criticality using feature attributes and spatial context available in vector data. Therefore, they cannot directly assess raster-only charts. Rule-based methods further restrict change classification to predefined rules. Vision-language models (VLMs) offer a complementary approach by interpreting rasterized charts and potentially leveraging learned knowledge beyond explicit rules. Given the safety-critical nature of maritime navigation, their ability to interpret nautical charts and assess navigational implications requires systematic evaluation. Existing geospatial benchmarks~\citep{li-2025-mapqa, pyo-2026-frieda, xu_2026_earthspatialbench} evaluate map understanding and spatial reasoning but omit nautical semantics, navigational restrictions, and safety considerations. Maritime vision benchmarks~\citep{bovcon-2021-mods, zust-2023-lars, kiefer-2023-macvi, wong-2026-marineeval} focus on on-board maritime imagery rather than the symbolic and geometric representations of nautical charts. ENC-Bench~\citep{cheng-2026-encbench} evaluates nautical-chart perception, spatial reasoning, and maritime decision-making, but not chart-change criticality or how underlying perceptual and spatial understanding inform such navigation-related decisions.


We introduce \benchname, a diagnostic benchmark of \benchsize questions derived from \noaasize NOAA ENCs. Motivated by navigational safety requirements, it evaluates four complementary categories: \textbf{ENC Perception}, \textbf{Spatial Reasoning} using neutral visual identifiers, \textbf{Joint Perception and Spatial Reasoning} grounded in nautical feature identities, and \textbf{Application} of navigational rules and safety considerations. Of these, we carefully design a progressive subset of \progressivesize questions that decomposes selected chart-change cases into linked questions spanning change detection, perception, spatial reasoning, joint reasoning, and criticality assessment. This enables case-level analysis of how lower-level perceptual and spatial evidence informs models' judgments of navigational significance.

Evaluations across 10 open-weight VLMs reveal limitations in fine-grained nautical feature recognition and spatial reasoning. Controlled evidence interventions improve change-criticality accuracy when decision-relevant perceptual and spatial information is provided. Our contributions are threefold:

\begin{itemize}
\item We introduce \benchname, a \benchsize-question benchmark derived from \noaasize NOAA ENCs, spanning 11 subtasks across four categories, with vector-derived reference answers and expert-reviewed change-criticality labels.

\item We develop a navigation standards-motivated diagnostic taxonomy that distinguishes nautical perception, semantic-independent spatial reasoning, entity-grounded spatial reasoning, and navigation-related decisions. A progressive subset of \progressivesize questions links these capabilities to change-criticality decisions within shared cases, enabling case-level evaluation.

\item We establish zero-shot baselines across 10 open-weight VLMs and examine how perceptual and spatial evidence affects change-criticality assessments through case-linked interventions.
\end{itemize}

\section{Related Work}
We position \benchname relative to prior work on geospatial and cartographic reasoning (Section~\ref{sec:geospatial-vqa}) and maritime and ENC understanding (Section~\ref{sec:enc-comprehension}).

\subsection{Geospatial and Cartographic Reasoning} \label{sec:geospatial-vqa}
Recent benchmarks evaluate geospatial reasoning across diverse maps. MapQA~\citep{li-2025-mapqa} examines spatial and multi-hop reasoning over geographic entities, while MapEval~\citep{dihan-2025-mapeval} evaluates textual, API-based, and visual map reasoning. EarthSpatialBench~\citep{xu_2026_earthspatialbench} and FRIEDA~\citep{pyo-2026-frieda} further evaluate topological, metric, and directional relationships in geospatial imagery and cartographic maps. However, these benchmarks do not address nautical symbology, navigational restrictions, or the safety implications of spatial relationships among nautical features. \benchname extends geospatial benchmarking to the maritime domain by evaluating domain-specific feature interpretation and navigational decisions alongside spatial reasoning.

\subsection{Maritime and ENC Understanding} \label{sec:enc-comprehension}
Maritime vision benchmarks~\citep{bovcon-2021-mods, zust-2023-lars, kiefer-2023-macvi, wong-2026-marineeval} primarily evaluate perception and scene understanding in onboard and aerial imagery rather than nautical charts. ENC-Bench~\citep{cheng-2026-encbench} evaluates nautical symbol recognition, spatial computation, and maritime decision-making, but does not assess chart-change criticality or systematically link intermediate capabilities to downstream navigational judgments.  \benchname addresses these gaps by evaluating visually grounded chart-change criticality and distinguishing semantic-independent from nautical-feature-grounded spatial reasoning. Case-linked questions and controlled evidence interventions further examine how perceptual and spatial understanding contribute to criticality judgments.

\section{Task Definitions} \label{sec:task-definition}
We design \benchname to evaluate the perceptual, spatial, and domain-specific reasoning capabilities required to interpret rasterized nautical charts and make navigation-related decisions. We ground this design in IHO charting specifications~\citep{iho-s52, iho-s57-appb, iho_s4_2021}, which define nautical feature representation and portrayal, and IMO navigational safety requirements~\citep{imo_ecdis, imo-msc232-2006}, which establish requirements for navigational decision-making. Reflecting these standards, we organize the benchmark into four complementary categories: nautical feature recognition (Section~\ref{sec:enc-perception}), semantic-independent spatial reasoning (Section~\ref{sec:spatial-reasoning}), feature-grounded spatial reasoning (Section~\ref{sec:joint}), and navigation-related decision-making (Section~\ref{sec:application}).


\subsection{ENC Perception} \label{sec:enc-perception}
ENC Perception evaluates whether models can recognize standardized nautical symbols and associate chart annotations with their corresponding features or locations. Nautical charts use internationally standardized symbology to represent feature types and characteristics, whose distinctions carry specific navigational implications~\citep{iho-s52, iho-s57-appb}. Textual annotations identify nautical features by name or designation and describe characteristics such as light flashing patterns, while numerical annotations convey quantitative information such as water depths~\citep{iho_s4_2021}. These visual elements convey the features and conditions relevant to navigational decisions and chart updates.


\begin{figure}[htbp]
    \centering

    \begin{subfigure}[b]{0.12\linewidth}
        \centering
        \parbox[c][1.5cm][c]{\linewidth}{%
            \centering
            \includegraphics[width=\linewidth]{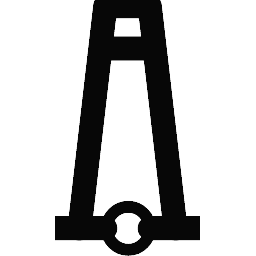}
        }
        \caption{}
        \label{fig:isolated-symbol-recognition}
    \end{subfigure}
    \hfill
    \begin{subfigure}[b]{0.25\linewidth}
        \centering
        \parbox[c][1.5cm][c]{\linewidth}{%
            \centering
            \includegraphics[width=\linewidth]{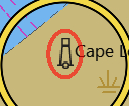}
        }
        \caption{}
        \label{fig:incontext-symbol-recognition}
    \end{subfigure}
    \hfill
    \begin{subfigure}[b]{0.60\linewidth}
        \centering
        \parbox[c][1.5cm][c]{\linewidth}{%
            \centering
            \includegraphics[width=\linewidth]{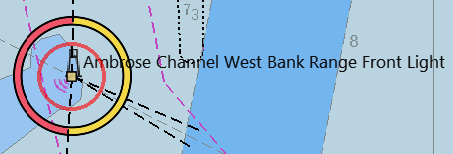}
        }
        \caption{}
        \label{fig:symbol-text-association}
    \end{subfigure}

    \vspace{-6pt}
    \caption{Overview of \textbf{ENC Perception} task.}
    \label{fig:enc-perception}
\end{figure}

\noindent\textbf{Isolated Symbol Recognition.}\quad
To assess models' foundational knowledge for nautical-chart interpretation, we present individual ENC symbols without surrounding chart context (Figure~\ref{fig:isolated-symbol-recognition}) and ask models to identify their precise meanings according to the IHO S-52 Presentation Library~\citep{iho-s52}. This isolates fine-grained symbol recognition from surrounding visual complexity.

\noindent\textbf{In-Context Symbol Recognition.}\quad
To evaluate precise symbol recognition under realistic chart complexity, we ask models to identify the meanings of marked nautical symbols within rendered ENCs containing neighboring features, labels, and overlapping graphical elements (Figure~\ref{fig:incontext-symbol-recognition}).

\noindent\textbf{Symbol-Text Association.}\quad
Recognizing chart text alone is insufficient, as models must also determine which nautical feature or location each annotation describes. Therefore, we evaluate whether models can associate annotations, such as geographic feature names, names of navigational features, and depth sounding values, with their corresponding symbols, areas, or positions (Figure~\ref{fig:symbol-text-association}). Correct associations ensure that feature identities and depth values are attributed to their intended locations, providing the necessary context for subsequent spatial reasoning and navigational judgments.

\subsection{Spatial Reasoning} \label{sec:spatial-reasoning}
Understanding the relative positions, distances, and interactions among nautical features is essential, as it can influence navigational decisions~\citep{imo-msc232-2006}. To evaluate these spatial capabilities within rasterized nautical charts, we assess topological, metric, and directional relations using neutral visual identifiers. By removing the requirement to recognize nautical features, this category provides a reference for examining the additional demands of nautical semantics in subsequent joint reasoning (Section~\ref{sec:joint}) and application (Section~\ref{sec:application}) tasks.

\begin{figure}[htbp]
    \centering

    \begin{subfigure}[b]{0.32\linewidth}
        \centering
        \includegraphics[height=2.0cm,keepaspectratio]{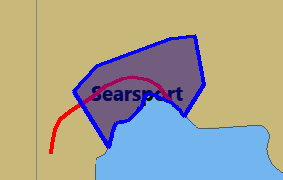}
        \caption{}
        \label{fig:topological}
    \end{subfigure}
    \hspace{0.075\linewidth}
    \begin{subfigure}[b]{0.32\linewidth}
        \centering
        \includegraphics[height=2.0cm,keepaspectratio]{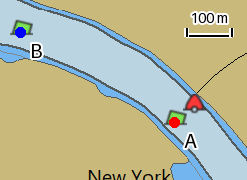}
        \caption{}
        \label{fig:metric}
    \end{subfigure}
    \hfill
    \begin{subfigure}[b]{0.23\linewidth}
        \centering
        \includegraphics[height=2.0cm,keepaspectratio]{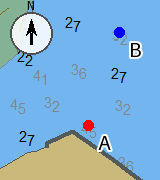}
        \caption{}
        \label{fig:directional}
    \end{subfigure}
    \vspace{-6pt}
    \caption{Overview of \textbf{Spatial Reasoning} task.}
    \label{fig:spatial-reasoning}
\end{figure}


\noindent\textbf{Topological Relations.}\quad
We evaluate \textit{within}, \textit{touches}, \textit{overlaps}, \textit{equals}, and \textit{crosses} across valid combinations (Appendix C) of point, line, and area geometries (Figure~\ref{fig:topological}). This task assesses whether models can distinguish spatial configurations such as containment, boundary contact, and crossing, which establish how charted features occupy and interact in space and provide a basis for interpreting navigational constraints~\citep{imo-msc232-2006}.

\noindent\textbf{Metric Relations.}\quad
We evaluate \textit{quantitative} distance estimation, measuring numerical distances between features, and \textit{qualitative} proximity relations, including identifying the nearest and farthest features (Figure~\ref{fig:metric}). This task assesses two complementary capabilities, essential for identifying nearby navigational hazards and evaluating clearance from charted features or boundaries~\citep{imo-msc232-2006}.

\noindent\textbf{Directional Relations.}\quad
We evaluate \textit{categorical} directional relations using cardinal and intercardinal directions, and \textit{numerical} bearings measured clockwise from true north~\citep{iho_s4_2021} (Figure~\ref{fig:directional}). This capability is essential for navigation decisions as it supports locating features relative to a reference position, interpreting route orientation, and determining the direction of navigational hazards~\citep{imo-msc232-2006}.

\subsection{Joint Perception and Spatial Reasoning} \label{sec:joint}
The navigational significance of a nautical feature depends on both its identity and its spatial relationships with surrounding features and routes. For example, identifying a wreck alone does not reveal its proximity to a navigational route, while determining proximity without recognizing the wreck provides limited navigational context. As ENC Perception (Section~\ref{sec:enc-perception}) and Spatial Reasoning (Section~\ref{sec:spatial-reasoning}) evaluate these capabilities independently, this category examines whether models can integrate nautical feature recognition with spatial reasoning.


\begin{figure}[htbp]
    \centering

    \begin{subfigure}[b]{0.48\linewidth}
    \centering
    \makebox[\linewidth][r]{%
        \includegraphics[width=0.85\linewidth]{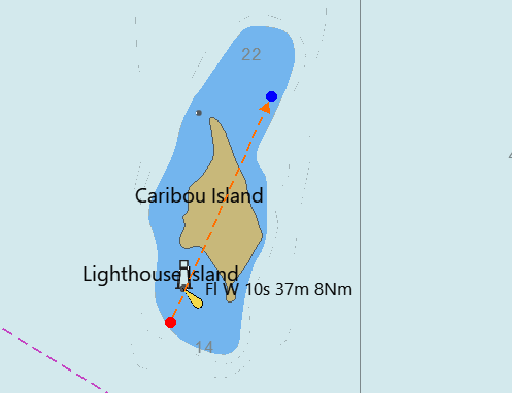}%
    }
    \caption{}
    \label{fig:path-relation}
\end{subfigure}%
\hspace{0.02\linewidth}%
\begin{subfigure}[b]{0.41\linewidth}
    \centering
    \makebox[\linewidth][l]{%
        \includegraphics[width=0.85\linewidth]{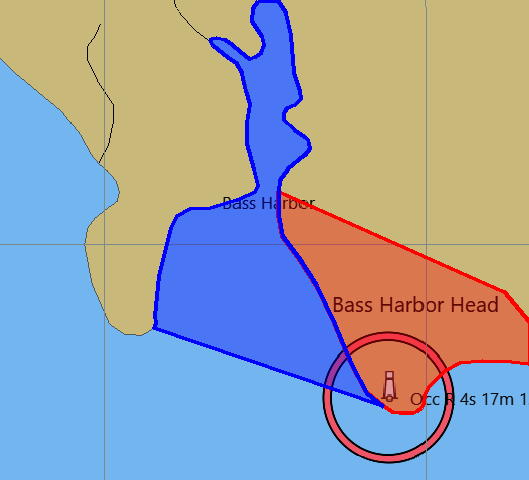}%
    }
    \caption{}
    \label{fig:spatial-relation}
\end{subfigure}

    \vspace{-6pt}
    \caption{Overview of \textbf{Joint Perception and Spatial Reasoning} task. Red and blue markers in (b) are
shown only for visualization.}
    \label{fig:spatial-reasoning}
\end{figure}

\noindent\textbf{Path-Based Reasoning.}\quad
To assess whether models can recognize nautical features and determine their interactions with a path, we overlay a straight-line path between specified endpoints on a rasterized nautical chart and ask which features it \textit{traverses}, \textit{crosses}, \textit{enters}, or \textit{exits} (Figure~\ref{fig:path-relation}). This task evaluates the joint capabilities needed to identify potential navigational hazards and interactions with regulated areas along a path.

\noindent\textbf{ENC Feature Spatial Relations.}\quad
Reasoning about spatial relationships among nautical features requires recognizing the features and determining their relative configurations. We evaluate the same spatial relations as in Spatial Reasoning (Section~\ref{sec:spatial-reasoning}), but reference features by their nautical identities rather than neutral visual identifiers (Figure~\ref{fig:spatial-relation}). This evaluates how feature recognition and spatial reasoning jointly support interpreting navigational conditions.

\subsection{Application} \label{sec:application}
Navigational decisions require applying operational rules, navigational restrictions, and safety requirements~\citep{imo-1972-convention, iho-s57-appb, imo-msc232-2006} to nautical features and their spatial relationships. We assess whether models possess and can apply this knowledge to evaluate vessel movements and chart-change criticality from rasterized nautical charts~\citep{iho_s4_2021}.

\begin{figure}[h]
\centering

\begin{subfigure}[c]{0.25\linewidth}
    \centering
    \includegraphics[width=0.95\linewidth]{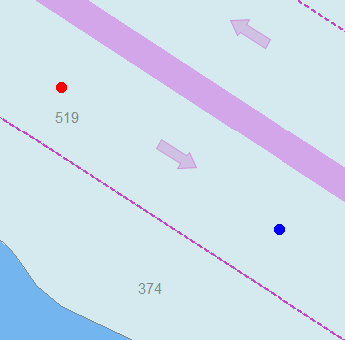}
    \caption{}
    \label{fig:travel-direction}
\end{subfigure}%
\hfill
\begin{subfigure}[c]{0.25\linewidth}
    \centering
    \includegraphics[width=0.95\linewidth]{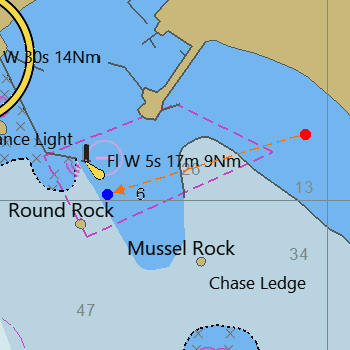}
    \caption{}
    \label{fig:travel-permission}
\end{subfigure}%
\hfill
\begin{subfigure}[c]{0.49\linewidth}
    \centering
    \includegraphics[width=0.45\linewidth]{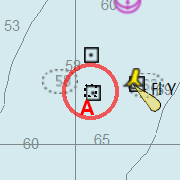}%
    \hspace{0.02\linewidth}%
    \includegraphics[width=0.45\linewidth]{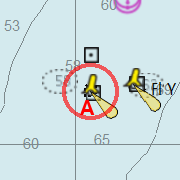}
    \caption{}
    \label{fig:change-criticality}
\end{subfigure}

\vspace{-6pt}
\caption{Overview of the \textbf{Application} task.}
\label{fig:application}

\end{figure}


\noindent\textbf{Travel Direction.}\quad
In traffic-separation schemes, maritime collision-prevention rules~\citep{imo-1972-convention} require vessels using a traffic lane to proceed in the general direction of its designated flow. We evaluate whether models can determine if a vessel traveling between specified start and end points complies with directional navigation requirements indicated by traffic lanes, fairways, and other nautical features (Figure~\ref{fig:travel-direction}). The task requires models not only to recognize the indicated traffic direction but also to apply navigational rules to determine whether the proposed movement is permitted.

\noindent\textbf{Travel Permission.}\quad
Nautical charts distinguish areas with varying operational constraints, where some prohibit passage while others indicate hazards without necessarily restricting access~\citep{iho-s57-appb}. We evaluate whether models can determine if travel through a specified region is permitted under the restrictions depicted in rasterized nautical charts (Figure~\ref{fig:travel-permission}). This task requires models to interpret applicable restrictions beyond simply recognizing regulated areas and their boundaries.


\noindent\textbf{Change Criticality Assessment.}\quad
We evaluate whether models can assess the navigational significance of changes between successive chart updates, including additions, deletions, positional shifts, and attribute modifications to nautical features (Figure~\ref{fig:change-criticality}). We classify a change as critical when the affected feature, its modified attributes, and surrounding spatial context satisfy predefined navigational safety criteria~\citep{iho_s4_2021, arndt-2026-enc}. Consequently, the same type of change may have different implications for navigational safety depending on its proximity to navigational routes, hazards, or restrictions. Models receive successive rasterized ENC charts along with structured descriptions of affected attributes and their previous and updated values. Since not all attribute changes are explicitly visible in chart renderings, these descriptions provide additional change information while requiring models to interpret the surrounding spatial and navigational context from the charts. For selected chart-change cases, we construct a \textbf{progressive} set that decomposes criticality assessments into linked questions covering change detection, perception, spatial reasoning, and their integration, leading to criticality judgments for the same cases. We use this case-linked design to examine how underlying capabilities relate to navigational judgments that require domain-specific knowledge (see Appendix~E for an example).

\section{\benchname} \label{sec:benchmark}

We design \benchname as a diagnostic benchmark for evaluating capabilities aimed at nautical chart understanding and downstream navigational decision-making. Constructed from \noaasize NOAA ENCs, the benchmark consists of \benchsize questions spanning the four capabilities defined in Section~\ref{sec:task-definition}. Of these, \progressivesize questions form a \textbf{progressive} set for case-linked analysis of how underlying capabilities contribute to change-criticality assessment. 


\subsection{Dataset Construction}\label{sec:dataset-construction}
\begin{figure*}[htbp]
    \centering
    \includegraphics[width=\linewidth]{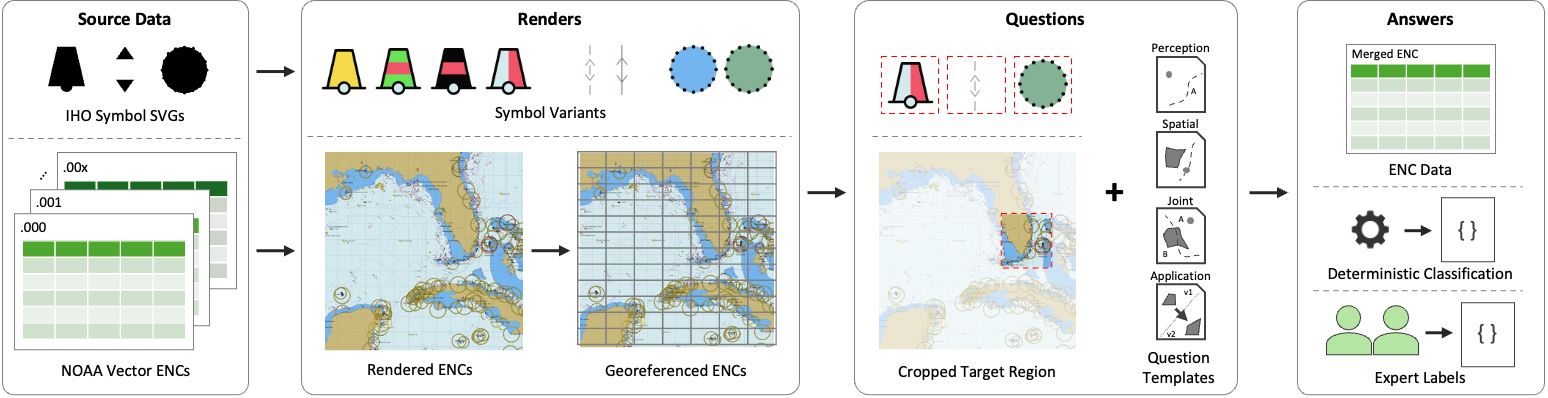}
    \caption{Overall dataset construction pipeline used to create \benchname.}
    \label{fig:pipeline}
    \vspace{-6pt}
\end{figure*}

Figure~\ref{fig:pipeline} illustrates our pipeline for creating the benchmark questions from NOAA ENC data~\citep{noaa_enc}. The pipeline consists of rendering, raster-vector alignment, visibility-aware selection, task-specific question generation, and answer construction.






\noindent\textbf{ENC Source Data.}\quad
Following the IHO S-57 standard~\citep{iho_s57_attribute_catalogue_2000, iho_s57_uoc_2024}, NOAA ENCs represent nautical features through standardized object classes, attributes, and point, line, or area geometries. Access to these underlying vector data allows us to deterministically extract feature identities and attributes and compute ground-truth spatial relationships, providing structured reference answers to questions generated from rasterized charts. The ENC source data covers diverse geographic regions and spans all six NOAA navigational scale ranges, from Overview to Berthing~\citep{noaa_rescheme}, representing varying levels of nautical detail.

\noindent\textbf{Symbol and Chart Rendering.}\quad
For Isolated Symbol Recognition, we reproduce individual nautical symbols with shapes and colors specified by the IHO S-52 Presentation Library~\citep{iho-s52}. For other tasks, we render ENCs using OpenCPN, an open-source navigation application~\citep{opencpn}. When available, we render charts at their compilation scale, which is the viewing scale intended by the chart producer. Otherwise, we select the largest map scale within the chart's assigned NOAA navigational scale range to maximize visible detail. To preserve visual resolution across large chart extents, we divide each chart into overlapping regions and render each as a chart piece. Appendix A provides additional rendering configurations.

\noindent\textbf{Raster-Vector Alignment.}\quad
We align each chart piece with its underlying vector data using geographic bounds, chart center coordinates, display scale, and image dimensions, then stitch the overlapping pieces into a continuous raster representation. The rendering parameters enable direct mapping of geographic coordinates in the source ENC data to pixel locations in the rendered charts. This alignment enables us to locate nautical features in rasterized charts by their geographic coordinates while retaining vector-derived information about their identities, attributes, and spatial relationships for question generation.

\noindent\textbf{Visibility-Aware Selection.}\quad
The visibility of nautical features and their attributes in rasterized nautical charts depends on IHO S-52 portrayal rules and rendering settings. For example, the \texttt{SCAMIN} attribute determines the scale at which a feature is no longer displayed as the chart is zoomed out~\citep{iho-s52, iho_s57_attribute_catalogue_2000}. For each chart, we record whether individual nautical features are visible at the selected rendering scale. We use these visibility records during question generation and answer construction to ensure that tasks requiring visual feature recognition reference features displayed in the rendered charts.



\subsection{Question Generation}\label{sec:question-generation}
We generate questions from aligned ENC raster charts and underlying vector data using task-specific sampling constraints to ensure valid spatial configurations and identifiable nautical features. For ENC Perception and Joint Perception and Spatial Reasoning tasks, we select questions such that all referenced nautical features are visible using the recorded visibility information. Spatial Reasoning does not require this visibility condition, as it references geometries through neutral visual identifiers. For all tasks, we verify that each question unambiguously identifies its referenced features or regions and has a uniquely determined reference answer. These constraints ensure questions contain the information needed to construct reliable answers.

Application questions combine standardized ENC attributes with domain-specific navigational rules. For Change Criticality Assessment, we compare successive ENC versions to identify additions, deletions, positional shifts, and attribute modifications. We pair raster charts covering the same geographic region across versions and provide structured JSON descriptions of changed attributes, including their previous and updated values, to communicate change information not explicitly visible in chart renderings. For the progressive set, we sample critical and non-critical changes involving comparable nautical feature classes and change types, then construct questions across the capability hierarchy using the same task-specific criteria described above.

\subsection{Answer Construction and Verification}\label{sec:answer-creation}
We construct semantic reference answers deterministically by converting encoded nautical feature classes, attributes, and symbol identifiers into standardized natural-language descriptions (e.g., \texttt{BOYPIL10} to red lateral pillar buoy). We compute spatial reference answers directly from the source ENC geometries. For change criticality, we assign labels using predefined classification rules~\citep{arndt-2026-enc} when they yield unambiguous outcomes, then refer unresolved cases to experts for review. This distinguishes rule-derived labels from those requiring domain-specific judgment. Each question–answer pair retains source-chart identifiers, feature references, geographic information, and relevant metadata, enabling traceability to the original ENC features and geometries for verification and reproducibility.

\paragraph{Expert Annotation of Change Criticality.} \label{sec:expert-annotation}
Nine domain experts with extensive experience in nautical-chart interpretation and navigational assessment participate in criticality annotation, including five senior reviewers. Each case is independently assessed by an initial annotator and a senior reviewer, without access to each other’s judgments. Disagreements are resolved in favor of the senior reviewer’s judgment.



\subsection{Dataset Statistics and Coverage}\label{sec:statistics}
\begin{figure}[h]
    \centering
    \includegraphics[width=0.9\linewidth]{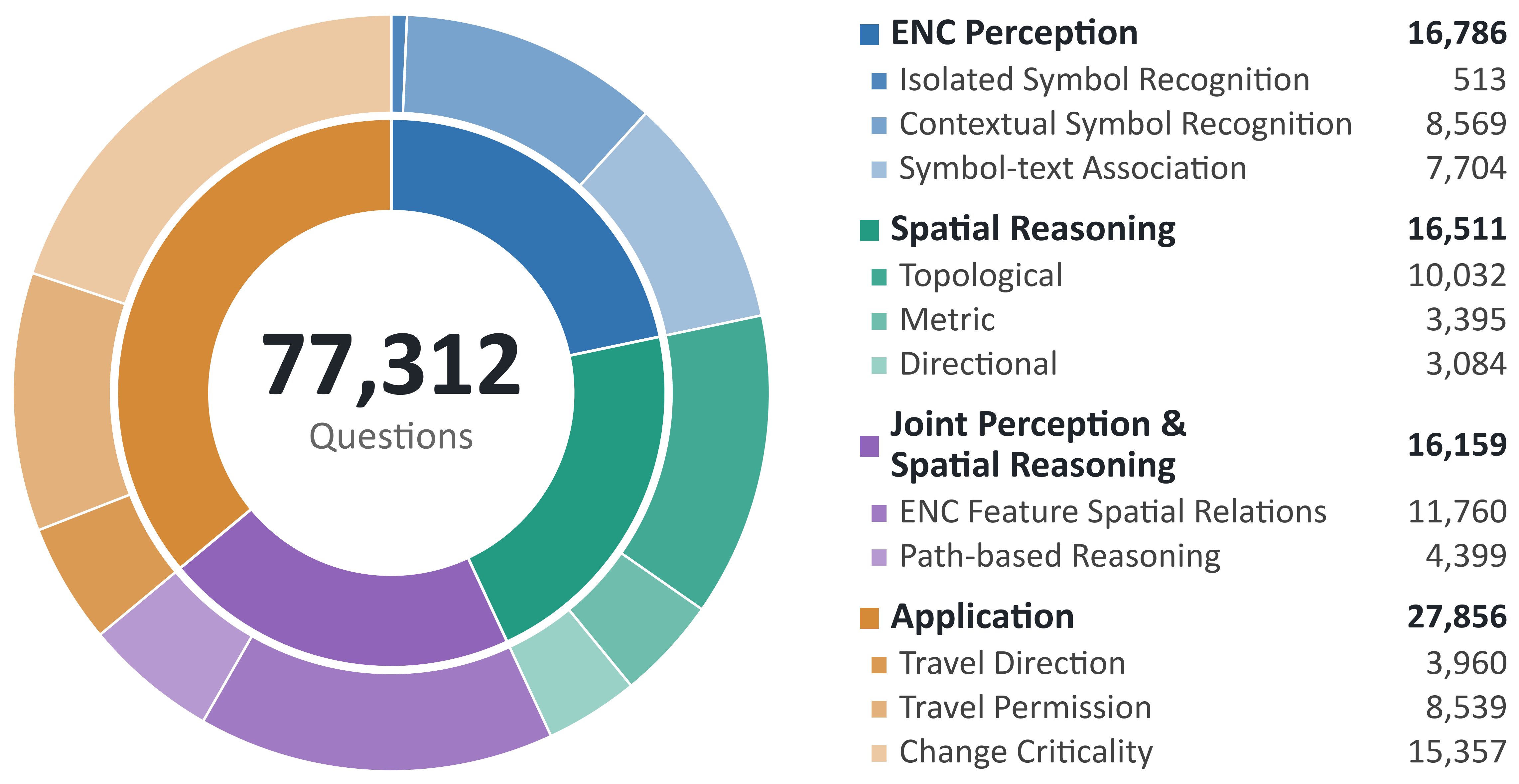}
    \vspace{-6pt}
    \caption{Dataset statistics of \benchname.}
    \vspace{-6pt}
    \label{fig:statistics}
\end{figure}

Figure~\ref{fig:statistics} illustrates the distribution of \benchname questions across the four capability categories and their respective subtasks, along with the total benchmark size. These counts include the \progressivesize-question progressive subset. Appendix D provides detailed distributions by geographic region and S-57 nautical feature class.


\section{Experiments}\label{sec:experiments}
We evaluate models on \benchname under a unified evaluation protocol. Models receive rasterized nautical charts and corresponding natural-language questions. For Change Criticality assessment, models additionally receive paired rasterized charts from successive ENC versions and structured JSON descriptions of the changes, including previous and updated attribute values. 
To evaluate chart understanding under the intended input conditions, we restrict models' access to the underlying ENC vector data, construction metadata, and external nautical-chart references. We also conduct complementary experiments using the progressive set to examine model performance under controlled evidence conditions.

\subsection{Baseline Models}\label{sec:baseline-models}
We evaluate 10 open-weight VLMs using fixed checkpoints and consistent inference settings for reproducible comparisons. To keep model scales comparable, we select variants with approximately 30B total parameters when available, or the largest available variant. All models use zero-shot prompting with temperature $0$, disabled sampling, standardized answer-format instructions, and a common maximum output length. Appendix F provides detailed inference configurations.

\subsection{Evaluation Metrics}\label{sec:evaluation-metric}
We first extract model answers from the designated \texttt{<answer>} tags and apply normalized exact matching for discrete textual responses, including binary yes-or-no answers, multiple-choice travel directions, counts, and categorical directions. For numerical distance questions, we compute mean absolute percentage error (MAPE) and consider predictions correct when their relative error is within 10\%~\citep{lewis-1982-mape}. For angular-bearing questions, we calculate the minimum circular angular difference between predicted and reference bearings and adopt a 10\textdegree\ correctness threshold~\citep{qian-2026-doa}. For Change Criticality assessment, we determine correctness by matching the critical-or-non-critical prediction against the reference label.

Since \benchname uses open-ended responses, semantically equivalent answers may use different expressions (e.g., \textit{within} vs. \textit{inside}) that fail exact matching. For textual responses that do not match after normalization, we employ an LLM-as-Judge~\citep{zheng-2023-llm-as-judge} to assess semantic correctness. The judge assigns a binary verdict based on semantic equivalence and answer specificity, rejecting responses that omit essential details or introduce contradictions. We use GPT-5.6 Sol~\citep{openai_gpt56_2026} as the judge model, with the evaluation prompt provided in Appendix~H.

\begin{table*}[t]
    \centering \small
    \renewcommand{\arraystretch}{1.15}
    \resizebox{\textwidth}{!}{%
    \begin{tabular}{l*{15}{c}}
        \toprule
        & \multicolumn{4}{c}{\textbf{ENC Perception}}
        & \multicolumn{4}{c}{\textbf{Spatial Reasoning}}
        & \multicolumn{3}{c}{\textbf{Joint}}
        & \multicolumn{4}{c}{\textbf{Application}} \\
        \cmidrule(lr){2-5}
        \cmidrule(lr){6-9}
        \cmidrule(lr){10-12}
        \cmidrule(lr){13-16}

        \textbf{Model}
        & \textbf{Sym} & \textbf{Ctx} & \textbf{Txt} & \textbf{Avg}
        & \textbf{Topo} & \textbf{Met} & \textbf{Dir} & \textbf{Avg}
        & \textbf{Path} & \textbf{Rel} & \textbf{Avg}
        & \textbf{TDir} & \textbf{TPer} & \textbf{Crit} & \textbf{Avg} \\
        \midrule
        Random Probability
        & -- & -- & -- & --
        & 31.9 & 25.1 & 9.0 & 26.2
        & -- & 12.3 & 8.8
        & 33.3 & 48.5 & 50.0 & 47.2 \\
        \midrule

        Qwen3.8-27B~\citep{team-2026-qwen38}
        & 0.2 & \textbf{7.3} & \textbf{60.8} & \textbf{31.6}
        & \textbf{48.2} & \textbf{71.0} & \textbf{83.1} & \textbf{59.4}
        & 67.2 & \textbf{49.7} & \textbf{54.5}
        & 60.4 & 67.6 & 59.7 & 62.2 \\
        
        LLaVA-OneVision2-8B~\citep{an-2026-llavaov2}
        & 0.0 & 3.4 & 50.0 & 24.7
        & 34.0 & 43.8 & 30.4 & 35.3
        & 54.8 & 28.2 & 35.5
        & 28.2 & 42.6 & 57.0 & 48.5 \\
        
        Qwen3.6-35B-A3B~\citep{team-2026-qwen-36}
        & \textbf{0.6} & 7.2 & \textbf{60.8} & \textbf{31.6}
        & 42.2 & 55.8 & 46.4 & 45.8
        & \textbf{69.8} & 37.7 & 46.4
        & 51.6 & 69.0 & 60.2 & 61.7 \\
        
        Ovis2.6-30B-A3B~\citep{lu-2025-ovis25}
        & 0.4 & 4.8 & 52.7 & 26.7
        & 40.4 & 40.3 & 37.6 & 39.9
        & 58.1 & 32.0 & 39.1
        & 54.5 & 48.3 & \textbf{62.7} & 57.1 \\
        
        Qwen3VL-30B-A3B-Instruct~\citep{bai-2025-qwen3vl}
        & 0.0 & 5.8 & 53.8 & 27.6
        & 38.0 & 45.3 & 38.5 & 39.6
        & 52.8 & 32.4 & 38.0
        & 33.2 & 71.3 & 56.2 & 57.6 \\
        
        InternVL3.5-30B-A3B~\citep{wang-2025-internvl35}
        & 0.0 & 3.7 & 48.6 & 24.2
        & 42.4 & 35.0 & 31.6 & 38.9
        & 46.2 & 33.1 & 36.7
        & 60.2 & 57.0 & 60.4 & 59.4 \\
        
        KimiVL-16B-A3B-Thinking~\citep{kimiteam-2025-kimivl}
        & 0.0 & 3.1 & 59.5 & 28.9
        & 40.9 & 39.6 & 36.4 & 39.8
        & 58.9 & 36.3 & 42.4
        & 24.0 & 54.5 & 57.5 & 51.8 \\
        
        KimiVL-16B-A3B-Instruct~\citep{kimiteam-2025-kimivl}
        & 0.0 & 2.6 & 58.3 & 28.1
        & 30.8 & 42.6 & 17.3 & 30.7
        & 63.2 & 21.6 & 32.9
        & 79.6 & 58.1 & 43.3 & 53.0 \\
        
        Idefics3-8B-LLaMA3~\citep{laurencon-2024-idefics3}
        & 0.0 & 2.7 & 37.9 & 18.8
        & 37.1 & 34.3 & 15.1 & 32.4
        & 48.9 & 22.8 & 29.9
        & 13.1 & 50.2 & 56.7 & 48.5 \\
        
        LLaVA-OneVision-7B~\citep{li-2024-llavaonevision}
        & 0.0 & 3.0 & 50.9 & 24.9
        & 40.8 & 38.5 & 15.8 & 35.7
        & 51.1 & 24.2 & 31.5
        & \textbf{80.2} & \textbf{79.6} & 54.1 & \textbf{65.6} \\
        
        \bottomrule
    \end{tabular}%
    }
    \vspace{-6pt}
    \caption{
        Zero-shot performance on \benchname (\%). Subtask abbreviations are grouped by category: ENC Perception (Sym: isolated symbol recognition; Ctx: in-context symbol recognition; Txt: symbol-text association), Spatial Reasoning (Topo: topological; Met: metric; Dir: directional), Joint Perception and Spatial Reasoning (Path: path-based; Rel: ENC feature spatial relations), and Application (TDir: travel direction; TPer: travel permission; Crit: change criticality). Avg indicates category-level accuracy. \textbf{Bold} denotes the best score in each column. Models are ordered by release date, newest first.
    }
    \label{tab:main_results}
    \vspace{-6pt}
\end{table*}

\section{Results}
We analyze baseline performance across the four capability categories (Section~\ref{sec:capability-analysis}), examine how perceptual and spatial evidence informs navigational judgments through the progressive set (Section~\ref{sec:progressive-analysis}), and evaluate the reliability of reference labels and automated scoring (Section~\ref{sec:reliability-analysis}).

\subsection{Capability-Level Performance} \label{sec:capability-analysis}
Table~\ref{tab:main_results} reports performance across the 11 subtasks grouped into four capability categories. Performance varies substantially across models and subtasks, with no single model consistently outperforming the others.

\noindent\textbf{ENC Perception.}\quad
Models achieve near-zero accuracy on Isolated and In-Context Symbol Recognition but perform substantially better on Symbol-Text Association. Qualitative inspection suggests that models often recognize broad nautical feature categories but fail to identify their precise characteristics. For example, models may correctly recognize a symbol as a \textit{buoy} but fail to distinguish characteristics such as whether it is a lateral buoy or a pillar buoy, even though these distinctions convey important navigational information. At the broader category level, accuracy increases by approximately 11 and 12 percentage points for Isolated and In-Context Symbol Recognition, respectively (Appendix G.1).

\noindent\textbf{Spatial and Joint Reasoning.}\quad
Across models, ENC Feature Spatial Relations (Joint) scores below the Spatial Reasoning category average. Unlike Spatial Reasoning, which uses neutral visual identifiers, this task requires identifying nautical features before evaluating their spatial relationships. The gap is consistent with additional feature-grounding demands.

\noindent\textbf{Application.}\quad
Change Criticality accuracy remains close to the 50\% random-guess baseline for most models, with the highest accuracy reaching 62.7\%. These results indicate that current VLMs remain unreliable for zero-shot assessment of the navigational significance of chart changes, despite access to chart imagery and structured change descriptions. To investigate whether these limitations reflect difficulties in visual perception, spatial reasoning, or applying nautical domain knowledge, we examine how providing task-relevant perceptual and spatial evidence affects criticality decisions using the progressive set.




\subsection{Progressive Case-Level Analysis} \label{sec:progressive-analysis}

\begin{table}[t]
    \centering \small
    \begin{tabular}{lcc}
         \toprule
         \textbf{Facts Provided} & \textbf{Accuracy (\%)} & \textbf{Gain (pp)} \\
         \midrule
         None & 57.5 & -- \\
         Perception & 65.3 & +7.8 \\
         Spatial & 63.9 & +6.4 \\
         Both & \textbf{81.0} & \textbf{+23.5} \\
         Irrelevant (\textit{Control}) & 49.7 & -7.8 \\
         \bottomrule
    \end{tabular}
    \vspace{-6pt}
    \caption{Change-criticality accuracy on the progressive subset under controlled evidence conditions. Perception and Spatial provide task-relevant facts individually, Both combines them, and Irrelevant (Control) provides a correct but decision-irrelevant fact. Gain (pp) denotes the percentage-point change relative to the no-evidence baseline (None).}
    \label{tab:progressive}
    \vspace{-6pt}
\end{table}


We evaluate change-criticality decisions on the progressive subset under five evidence conditions: no additional facts, perceptual facts only, spatial facts only, both, and a control providing a correct but decision-irrelevant fact. As shown in Table~\ref{tab:progressive}, providing both perceptual and spatial evidence increases accuracy by 23.5 percentage points, exceeding the gains from either alone (7.8 and 6.4 points, respectively). In contrast, irrelevant evidence reduces accuracy by 7.8 points, suggesting that improvements depend on the relevance of the supplied information rather than additional context.

These findings suggest that current VLMs can use nautical and spatial information for change-criticality judgments but struggle to do so without explicit intermediate evidence. The larger improvement from combining perceptual and spatial facts highlights the importance of integrating nautical feature identification with spatial context, rather than relying on either capability independently. Yet, accuracy remains at 81.0\% even when both types of evidence are provided, suggesting unresolved difficulties in applying nautical domain knowledge or integrating the supplied facts into navigational judgments. The decline in accuracy when given correct but irrelevant facts suggests that models may struggle to distinguish information relevant to a criticality decision from information that should not affect it.

\subsection{Reliability of Evaluation and Ground Truth} \label{sec:reliability-analysis}
\noindent\textbf{Expert Annotation Reliability.}\quad
Across 8,646 change-criticality cases requiring expert review, initial and senior reviewers agreed on 8,118 cases (93.9\%), with Cohen's $\kappa=0.862$, indicating strong inter-annotator agreement.

\noindent\textbf{LLM-as-Judge Reliability.}\quad
We select 99 model responses requiring semantic judgment across task categories and VLMs used in evaluation. Two human annotators independently assess each response against its reference answer using the same evaluation criteria as the LLM-as-Judge. The annotators agree on 93 responses (93.9\%), with Cohen's $\kappa = 0.840$. Among these 93 responses, GPT-5.6 Sol agrees with the human consensus on 87 (93.5\%, $\kappa = 0.827$). These results indicate strong human-to-human and human-to-judge agreement, supporting the reliability of our automated semantic evaluation. Appendix H.1 presents representative cases of disagreement between the LLM judge and human consensus.

\section{Conclusion}
We introduce \benchname, a diagnostic benchmark comprising \benchsize questions derived from \noaasize NOAA ENCs, spanning nautical feature perception, spatial reasoning, their joint application, and navigational decision-making. Evaluating 10 open-weight VLMs reveals substantial limitations in fine-grained nautical feature recognition and feature-grounded spatial reasoning, with change-criticality performance remaining near chance for most models. Our progressive evaluation further shows that supplying both perceptual and spatial evidence improves criticality accuracy by 23.5 percentage points, while irrelevant evidence reduces performance by 7.8 points. These findings suggest that current VLMs can leverage domain-relevant information for navigational judgments but struggle to reliably extract, integrate, and distinguish the evidence needed for those decisions. Moreover, remaining errors even with explicit evidence indicate unresolved challenges in applying nautical knowledge and reasoning over chart changes. By connecting intermediate capabilities to downstream decisions, \benchname provides a structured framework for diagnosing these limitations and advancing more reliable nautical-chart understanding.

\section*{LLM Usage Statement}
We used LLMs (GPT-6) to assist with manuscript preparation, including language refinement, structural revisions, and LaTeX formatting. We also used LLMs to review and refine author-produced content, including related work, code implementation, and debugging. The authors reviewed and verified all AI-assisted content and retain full responsibility for the manuscript's scientific claims, experimental results, and conclusions.

Additionally, we used GPT-5.6 Sol as an LLM-as-Judge to assess the semantic correctness of free-form model responses that could not be evaluated through deterministic matching. We assessed the reliability of these judgments through independent human annotation on a subset of responses and agreement analysis. Section~\ref{sec:answer-creation} and Appendix~H describe the evaluation procedure and judge prompt, respectively, and Section~\ref{sec:reliability-analysis} reports the reliability results.



\section*{Acknowledgments}
We acknowledge that this manuscript has been authored by UT-Battelle, LLC under Contract No. DE-AC05-00OR22725 with the U.S. Department of Energy. The United States Government retains and the publisher, by accepting the article for publication, acknowledges that the United States Government retains a non-exclusive, paid-up, irrevocable, world-wide license to publish or reproduce the published form of this manuscript, or allow others to do so, for United States Government purposes. DOE will provide public access to these results of federally sponsored research in accordance with the DOE Public Access Plan (http://energy.gov/downloads/doe-public-access-plan). Research sponsored by the Laboratory Directed Research and Development Program of Oak Ridge National Laboratory, managed by UT-Battelle, LLC, for the U. S. Department of Energy.

This research was supported in part by an appointment to the Oak Ridge National Laboratory GRO Program, sponsored by the U.S. Department of Energy and administered by the Oak Ridge Institute for Science and Education.

Additional support for this work was provided through computational resources from the Neural Signals and Computations lab via access to their computational server that is funded in part by NSF CAREER Award 2340338 (PI Dr. Adam Charles) and NIH NIBIB award R01EB037653 (PI Dr. Adam Charles).

{
    \small
    \bibliographystyle{ieeenat_fullname}
    \bibliography{main}
}


\newpage
\newpage
\appendix

\section{OpenCPN Settings} \label{app:opencpn-setting}

We use OpenCPN 5.14.0 to render NOAA ENCs. We downloaded the ENC cells from the NOAA ENC distribution webpage\footnote{\url{https://charts.noaa.gov/ENCs/ENCs.shtml}} on June 1, 2026, and use this collection as a fixed snapshot throughout benchmark construction. Table~\ref{tab:opencpn_setting} summarizes the OpenCPN portrayal settings applied consistently across all rendered samples.

\begin{table}[h]
\centering
\begin{tabular}{ll}
\toprule
\textbf{Configuration} & \textbf{Setting} \\
\midrule
OpenCPN version                & 5.14.0 \\
ENC display category           & All \\
Symbol style                   & Paper chart \\
Boundary style                 & Plain boundaries \\
S-57 text                      & Enabled \\
Soundings                      & Enabled \\
Metadata objects               & Enabled \\
SCAMIN                         & Enabled \\
Text decluttering              & Enabled \\
Light descriptions             & Enabled \\
Extended light sectors         & Enabled \\
National-language text         & Enabled \\
Color scheme                   & Day \\
Object scaling                 & Default \\
Sounding scaling               & Default \\
ENC text scaling               & Default \\
OpenGL rendering               & Enabled \\
\bottomrule
\end{tabular}
\caption{OpenCPN configuration used to render ENC images in \benchname.}
\label{tab:opencpn_setting}
\end{table}

\begin{figure*}[htbp]
    \centering
    \begin{subfigure}[b]{0.495\linewidth}
        \centering
        \includegraphics[width=\linewidth]{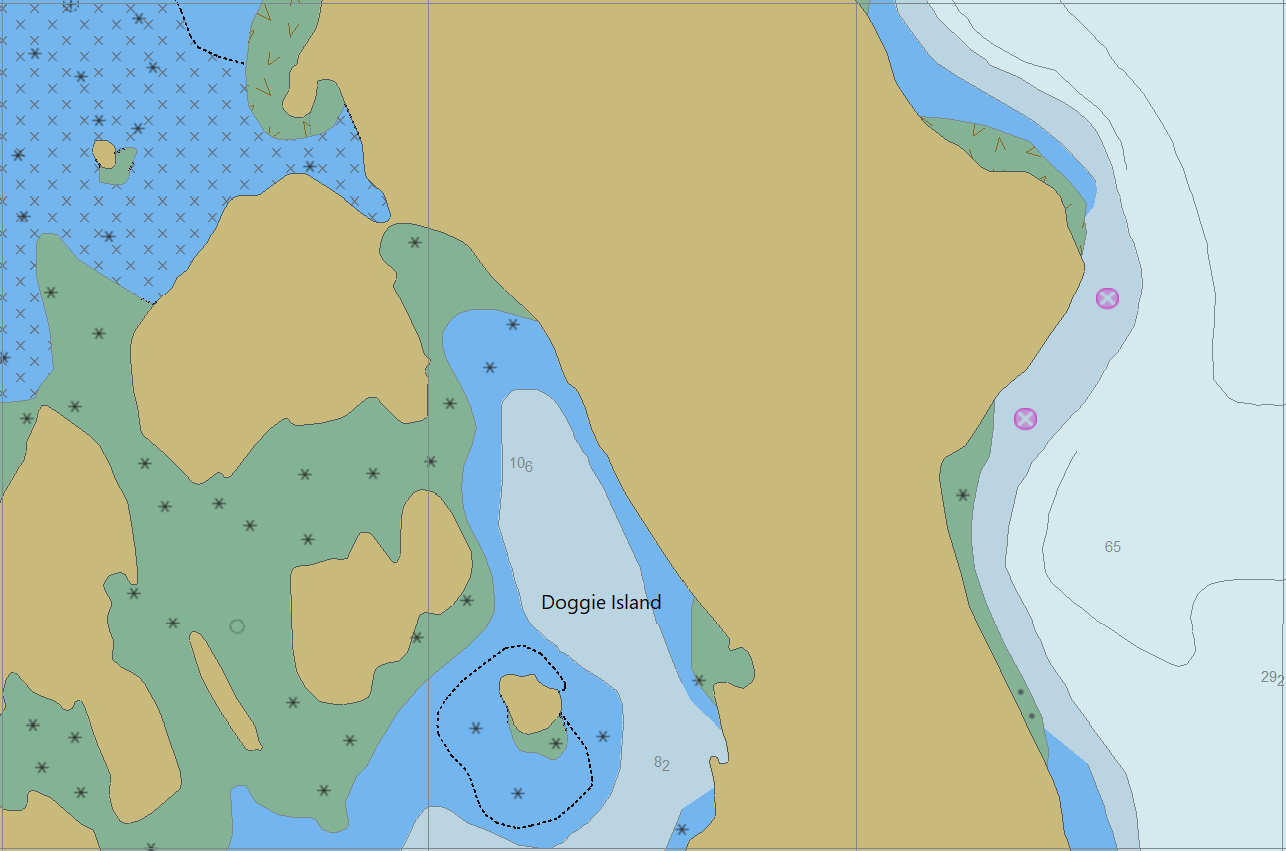}
        \caption{Render on Windows}
        \label{fig:render-windows}
    \end{subfigure}
    \hfill
    \begin{subfigure}[b]{0.495\linewidth}
        \centering
        \includegraphics[width=\linewidth]{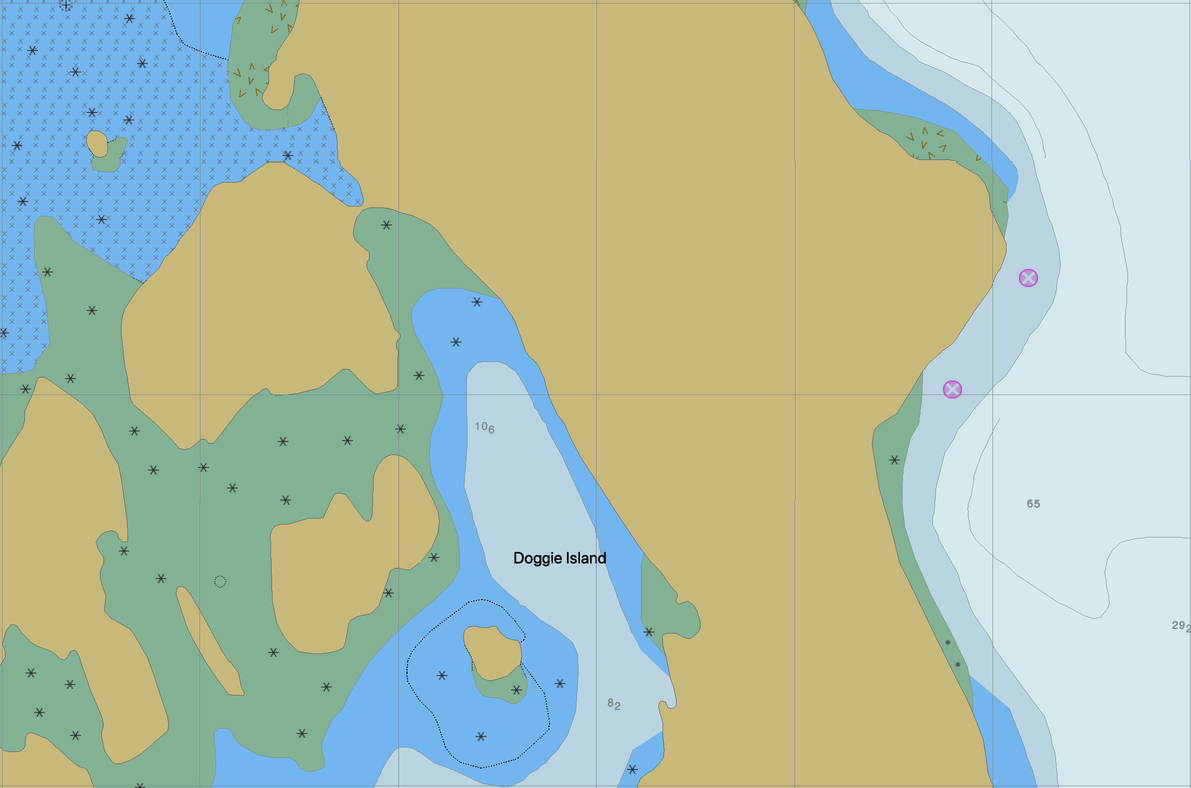}
        \caption{Render on macOS}
        \label{fig:render-mac}
    \end{subfigure}
    \caption{Comparison of ENC renderings generated on Windows and macOS using identical OpenCPN settings.}
    \label{fig:render-main}
\end{figure*}

Although our automated rendering pipeline supports both Windows (Figure~\ref{fig:render-windows}) and macOS (Figure~\ref {fig:render-mac}), we observed differences in the default fonts used for ENC text when rendering the same charts with identical OpenCPN settings. To eliminate platform-dependent font variations as a potential source of inconsistency, we render all benchmark images on Windows using the configuration in Table~\ref{tab:opencpn_setting}.


\section{Render Quality Validation}
We manually validate raster-vector alignment through 10 full-chart overlays and five feature-level crop tests, selecting charts across different geographic regions and rendering scales. Full-chart overlays assess the spatial correspondence between rendered nautical features and their source ENC geometries, while crop tests verify that geographic regions map to their expected locations in the image. These checks provide qualitative evidence of alignment consistency across the inspected samples.

\subsection{Full Chart Overlay} \label{app:full-chart-overlay}
\begin{figure}[htbp]
    \centering
    \includegraphics[width=\linewidth]{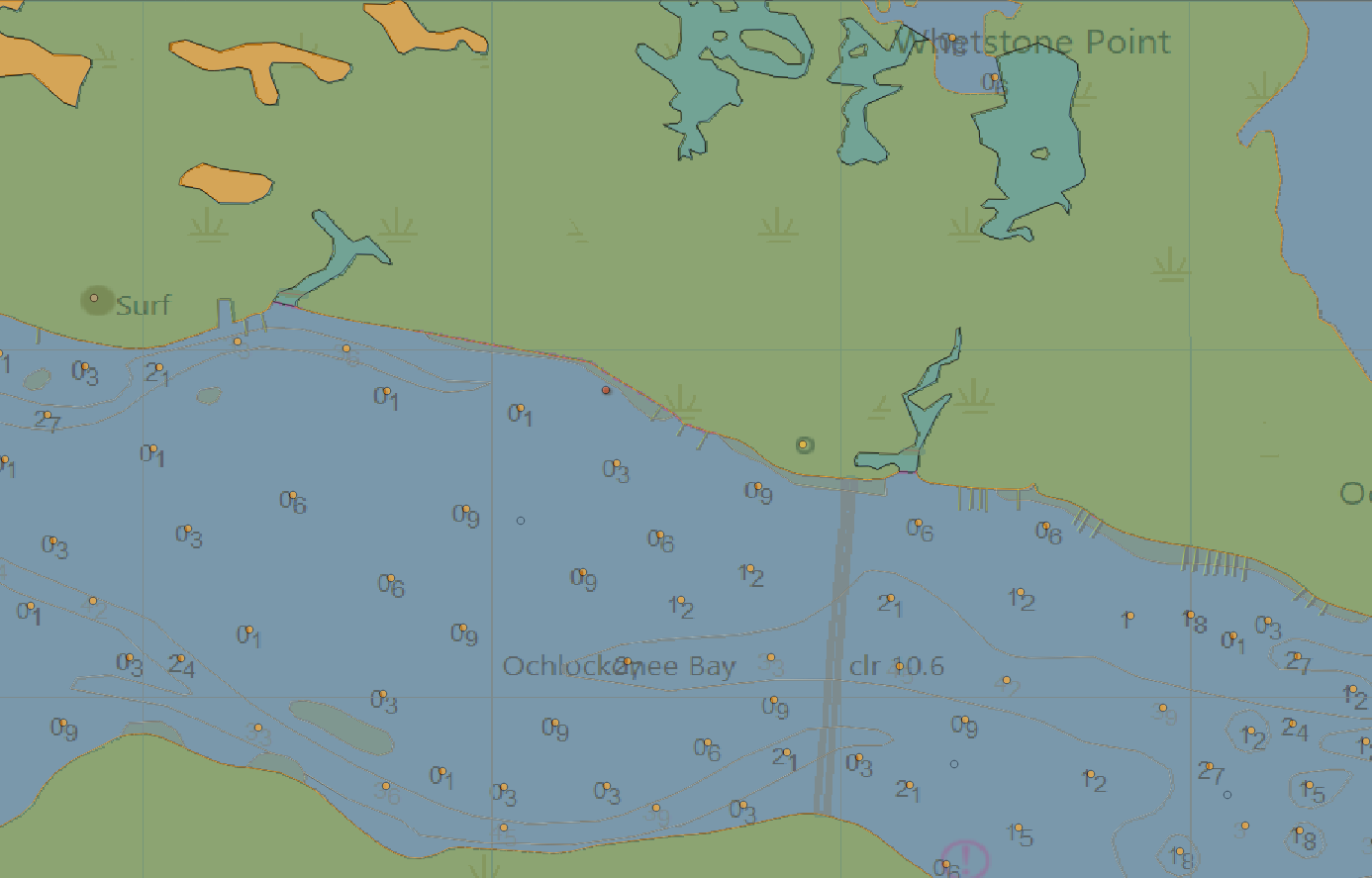}
    \caption{Magnified view of the US4FL1WK raster-vector overlay, illustrating the alignment of rendered coastlines and polygon boundaries with their source geometries.}
    \label{fig:overlay-zoom}
\end{figure}

\begin{figure*}[thbp]
    \centering
    \includegraphics[width=\linewidth]{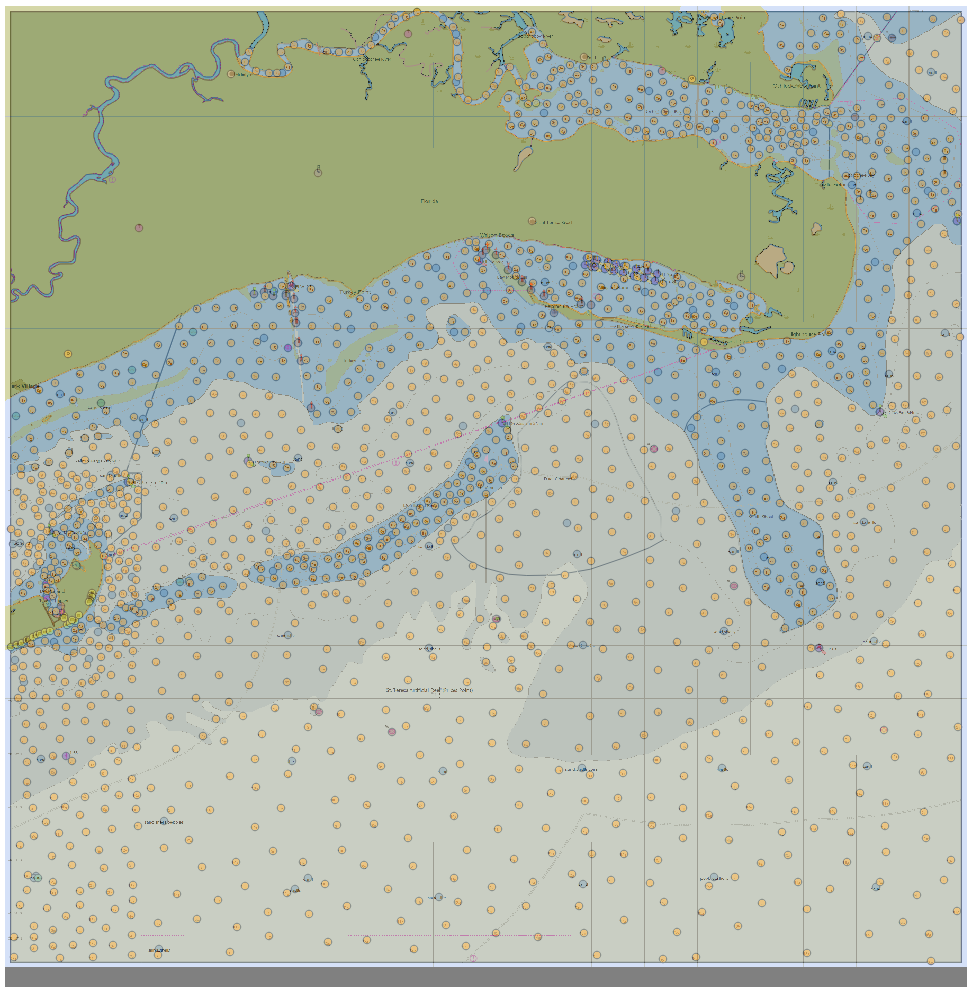}
    \caption{Full-chart raster-vector overlay for US4FL1WK at 1:45,000 scale, showing the spatial alignment between the rendered chart and source ENC geometries.}
    \label{fig:overlay-full}
\end{figure*}

We load the source ENC vector geometries and corresponding georeferenced raster charts into QGIS\footnote{\url{https://qgis.org/}} to visually inspect their alignment. Figure~\ref{fig:overlay-full} shows the full-chart overlay for US4FL1WK rendered at a scale of 1:45,000\footnote{The supplementary \texttt{render\_validation} folder contains the original US4FL1WK ENC vector data and its rendered chart at the evaluated scale.} while Figure~\ref{fig:overlay-zoom} presents a magnified region. Both views show consistent alignment between rendered coastlines, polygon boundaries, and their corresponding source geometries.

Minor differences may occur along feature boundaries because continuous vector geometries are represented on a discrete pixel grid during rasterization. Such pixel-level discrepancies do not necessarily indicate errors in geographic alignment.

\subsection{Crop test} \label{app:crop-test}
\begin{figure}[htbp]
    \centering
    \begin{subfigure}[b]{0.495\linewidth}
        \centering
        \includegraphics[width=\linewidth]{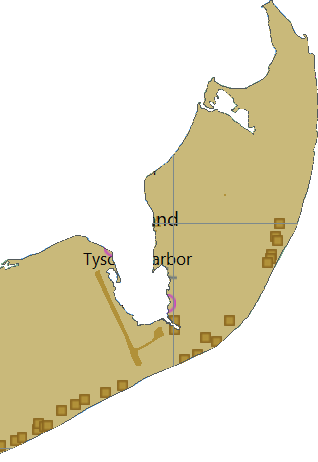}
        \caption{Land Area (LNDARE)}
        \label{fig:lndare-crop}
    \end{subfigure}
    \hfill
    \begin{subfigure}[b]{0.495\linewidth}
        \centering
        \includegraphics[width=\linewidth]{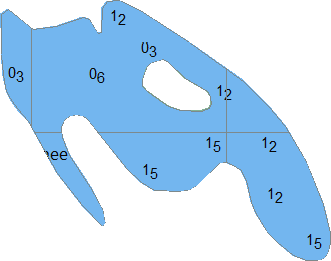}
        \caption{Depth Area (DEPARE)}
        \label{fig:depare-crop}
    \end{subfigure}
    \caption{Representative feature-level crops extracted using the geographic extents of source ENC polygons. Both crops contain their corresponding portrayed regions at the expected image locations.}
    \label{fig:crop-test}
\end{figure}

We further validate geographic-to-pixel mapping by extracting image crops from the known geographic extents of source ENC polygons, including land areas (\texttt{LNDARE}) and depth areas (\texttt{DEPARE}). Figure~\ref{fig:crop-test} shows representative examples in which the extracted crops contain the corresponding portrayed features at their expected image locations, providing localized evidence of raster-vector alignment.

\section{Spatial Reasoning Geometry Constraints}

We restrict the geometry combinations used in Spatial Reasoning questions to ensure that each spatial relationship is well-defined and consistent with its question formulation. Table~\ref{tab:spatial_geometry_combinations} lists the permitted pairings of points, lines, and polygons. Topological relations follow predicate-specific constraints, while quantitative distance and directional questions use point pairs. Qualitative distance questions permit combinations of all three geometry types.

\begin{table*}[t]
\centering
\begin{tabular}{lll}
\toprule
\textbf{Category} & \textbf{Question Type} & \textbf{Valid Geometry Combinations} \\
\midrule

\multirow{5}{*}{Topological}
& Within  & Point-Polygon, Line-Polygon, Polygon-Polygon \\
& Overlap & Polygon-Polygon \\
& Touch   & Point-Line, Point-Polygon, Line-Line, Line-Polygon, Polygon-Polygon \\
& Cross   & Line-Line, Line-Polygon \\
& Equal   & Point-Point, Line-Line, Polygon-Polygon \\

\midrule

\multirow{2}{*}{Distance}
& Quantitative & Point-Point \\
& Qualitative  & Point-Point, Point-Line, Point-Polygon, Line-Point, Line-Line, \\
&              & Line-Polygon, Polygon-Point, Polygon-Line, Polygon-Polygon \\

\midrule

\multirow{2}{*}{Direction}
& 8-Cardinal    & Point-Point \\
& Exact Bearing & Point-Point \\

\bottomrule
\end{tabular}
\caption{Permitted geometry pairings for Spatial Reasoning question generation. Pairings are ordered, with the first geometry serving as the subject and the second as the reference (e.g., Point-Polygon for \textit{Within} denotes a point within a polygon area).}
\label{tab:spatial_geometry_combinations}
\end{table*}
\section{\benchname Coverage}
We characterize \benchname's coverage by the geographic extent of its source ENCs and the diversity of S-57 object classes represented in the benchmark.

\subsection{Geographic Coverage} \label{app:geographic-coverage}
\begin{figure}[htbp]
    \centering
    \includegraphics[width=\linewidth]{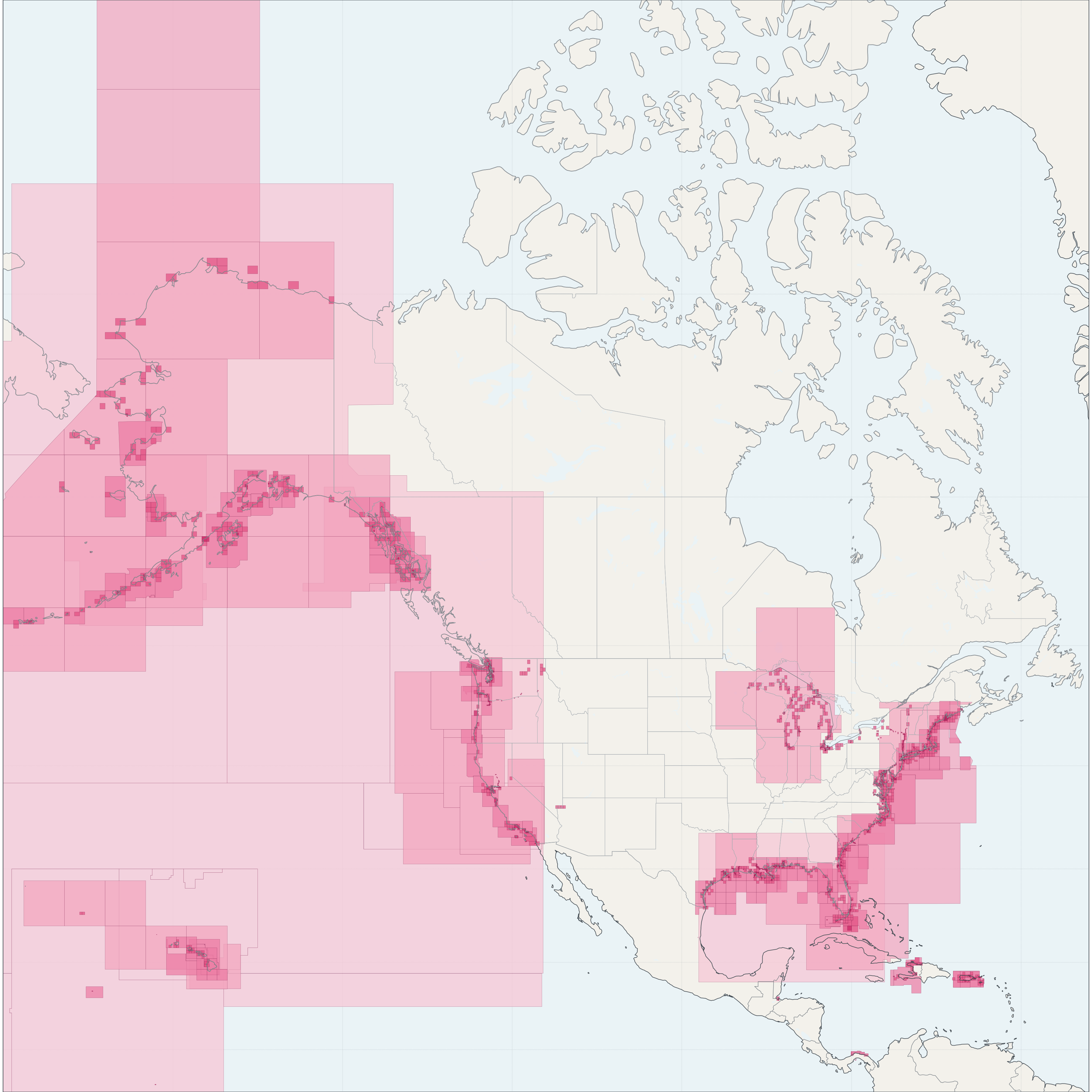}
    \caption{Geographic coverage of the NOAA ENC cells used to construct \benchname. The benchmark spans a broad range of U.S. coastal and maritime regions, including large-scale coastal waters and more localized harbor and berthing areas.}
    \label{fig:geographic-coverage}
\end{figure}

We construct \benchname using all available NOAA ENC cells from our collected snapshot. Figure~\ref{fig:geographic-coverage} illustrates the geographic distribution of these source charts, spanning diverse maritime environments and chart scales.

\subsection{S-57 Object Coverage} \label{app:object-coverage}
\begin{figure}[htbp]
    \centering
    \includegraphics[width=\linewidth]{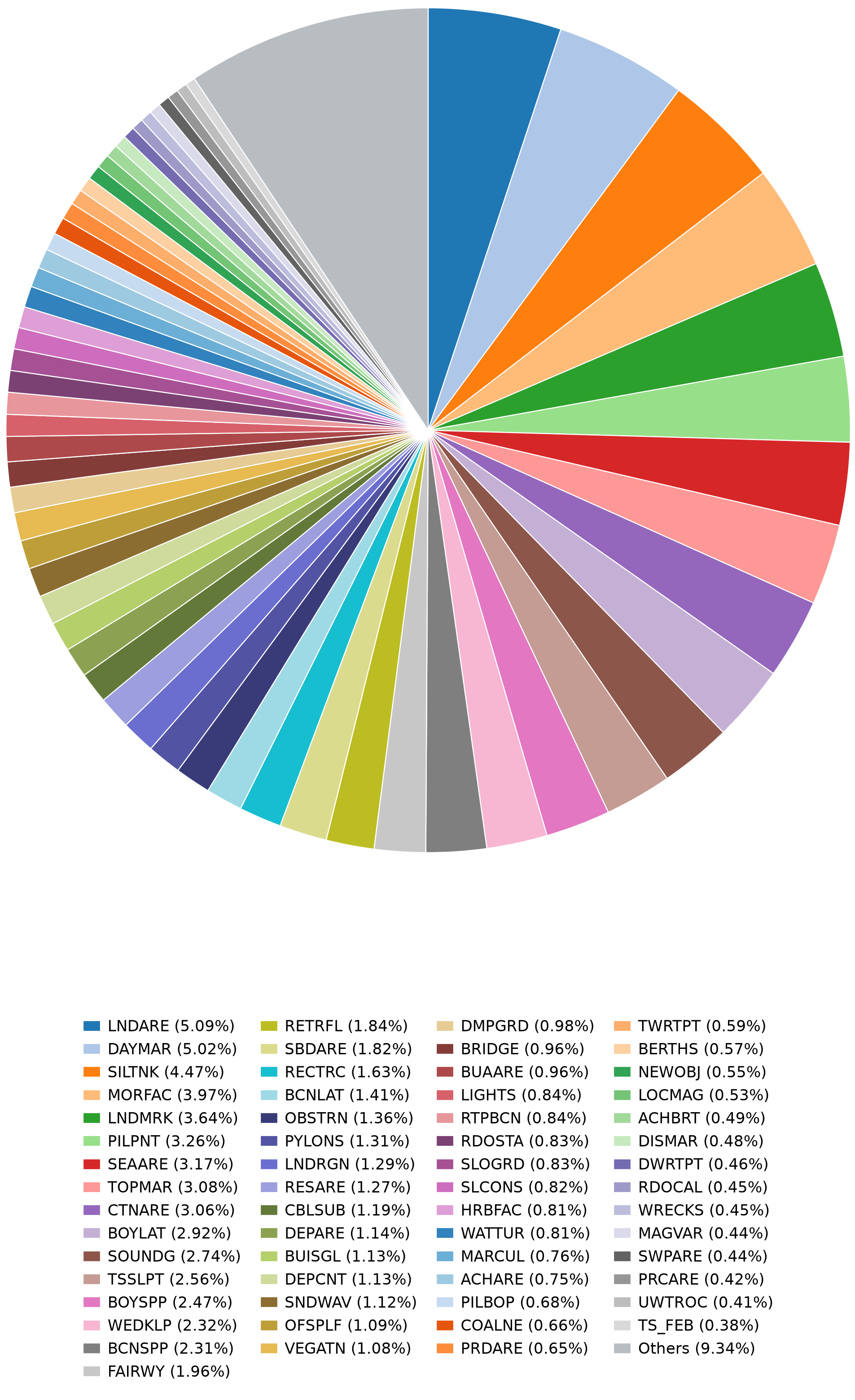}
    \caption{Distribution of S-57 object classes represented in \benchname. Less frequent classes are grouped under \textit{Other}.}
    \label{fig:pie-chart}
\end{figure}

We examine the distribution of S-57 object classes represented in \benchname to assess the diversity of nautical features included in the benchmark. During question generation, we balance entity sampling across object classes to reduce overrepresentation of frequently occurring features and encourage broader coverage. Figure~\ref{fig:pie-chart} presents the most frequent S-57 object classes, grouping less frequent classes into an \textit{Other} category.

\section{Progressive Set Example}
\begin{figure*}[t]
    \centering
    \begin{subfigure}[b]{0.47\textwidth}
        \centering
        \includegraphics[width=\linewidth]{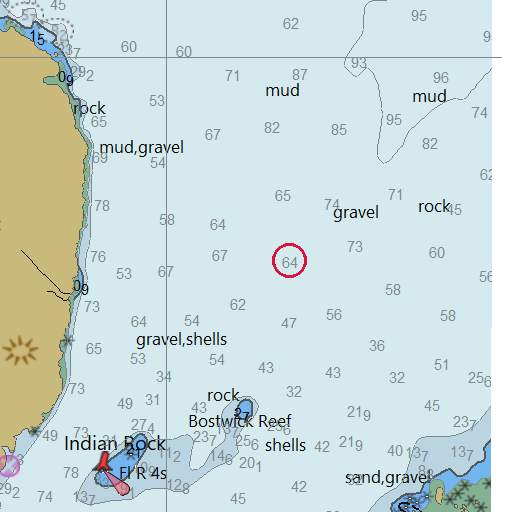}
        \caption{Original chart with highlighted location}
        \label{fig:progressive-old-marked}
    \end{subfigure}
    \hfill
    \begin{subfigure}[b]{0.47\textwidth}
        \centering
        \includegraphics[width=\linewidth]{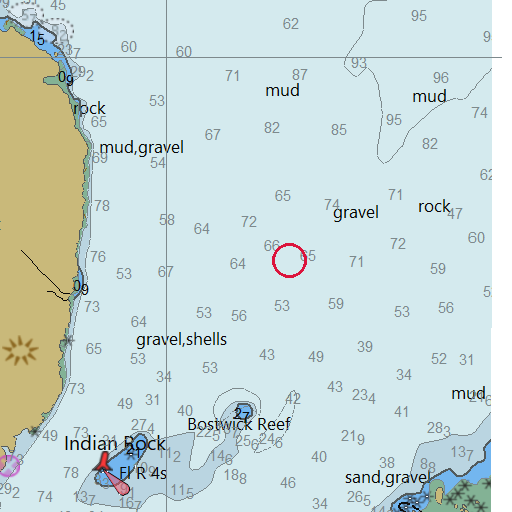}
        \caption{Updated chart with highlighted location}
        \label{fig:progressive-new-marked}
    \end{subfigure}

    \vspace{4pt}

    \begin{subfigure}[b]{0.45\textwidth}
        \centering
        \includegraphics[width=\linewidth]{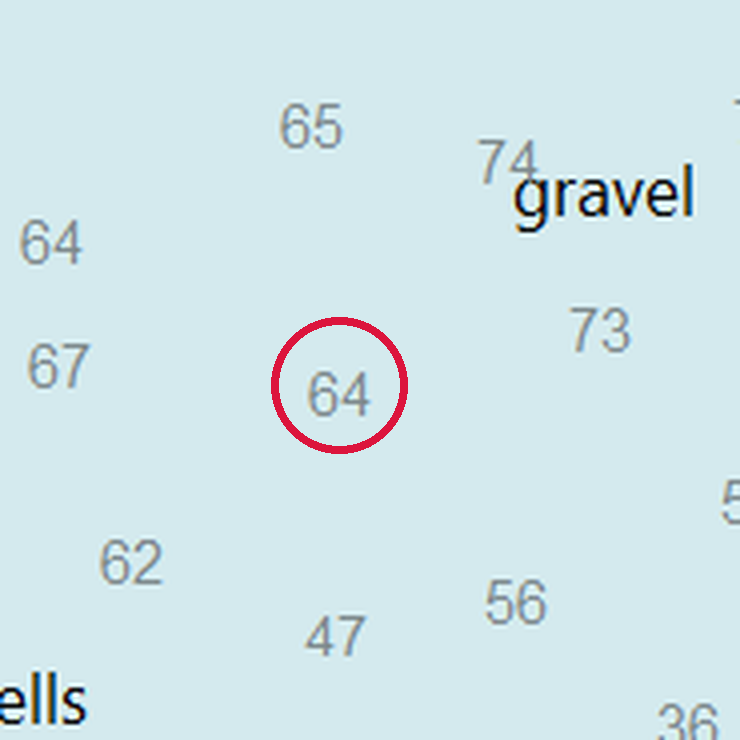}
        \caption{Zoomed-in view on sounding for perception}
        \label{fig:progressive-perception}
    \end{subfigure}
    \hfill
    \begin{subfigure}[b]{0.47\textwidth}
        \centering
        \includegraphics[width=\linewidth]{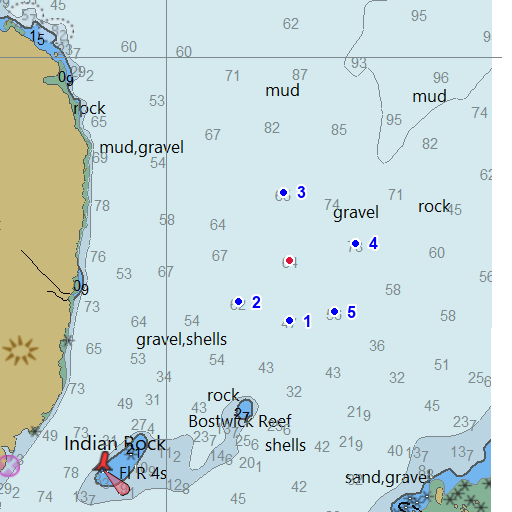}
        \caption{Neutral markers for spatial reasoning}
        \label{fig:progressive-spatial}
    \end{subfigure}

    \caption{Representative chart-change case from the progressive subset. The original and updated ENC renderings (a-b) provide context for Change Criticality Assessment. The magnified sounding (c) supports feature identification and depth interpretation, while the neutral-marker view (d) supports spatial reasoning.}
    \label{fig:progressive-example}
\end{figure*}

Figure~\ref{fig:progressive-example} illustrates a representative case from the progressive subset, linking intermediate chart-understanding questions to a downstream Change Criticality Assessment. The example includes paired chart versions, a zoomed view of the referenced sounding, and a spatial reasoning image with neutral markers. Table~\ref{tab:progressive-example} presents the corresponding questions and reference answers across four capability categories.

\begin{table*}[t]
    \centering
    \small
    \renewcommand{\arraystretch}{1.2}
    \begin{tabular}{p{0.18\textwidth} p{0.61\textwidth} p{0.10\textwidth}}
        \toprule
        \textbf{Category} & \textbf{Question} & \textbf{Answer} \\
        \midrule

        \multirow{2}{*}{ENC Perception}
        & What type of nautical feature is circled in red?
        & Sounding \\
        & What depth, in meters, is represented by the circled feature?
        & 64 m \\

        \midrule
        Spatial Reasoning
        & Which numbered blue marker is closest to the red marker? Answer with the corresponding number.
        & 1 \\

        \midrule
        Joint
        & Among the five soundings nearest to the circled feature, what is the shoalest charted depth in meters?
        & 47 m \\

        \midrule
        Application
        & The first image is the old chart and the second image is the new chart. Is the change at the circled location critical to navigational safety?
        & Yes \\

        \bottomrule
    \end{tabular}
    \caption{Representative questions and reference answers from the progressive subset based on images in Figure~\ref{fig:progressive-example}.}
    \label{tab:progressive-example}
\end{table*}

The Perception questions assess recognition of sounding values and interpretation of the text associated with the sounding. The Spatial Reasoning question evaluates proximity using neutral identifiers, without requiring recognition of nautical features. Joint Perception and Spatial Reasoning combine feature identification with spatial reasoning of nearby soundings. Change criticality uses the information from Joint Reasoning to determine whether the chart modification is navigationally significant under the applicable rule. These linked questions illustrate how the progressive subset enables case-level examination of intermediate capabilities relevant to downstream navigational decisions.
\section{Inference Setting}
We conduct Isolated Symbol Recognition experiments on an NVIDIA GB10, and all remaining experiments use four NVIDIA RTX 6000 Ada GPUs and two NVIDIA A40 GPUs. For experiments on the RTX 6000 Ada and A40 GPUs, we use consistent inference settings with BitsAndBytes 4-bit quantization, FlashAttention-2, and a batch size of 8. We evaluate all models zero-shot with deterministic decoding (temperature $0$ and sampling disabled), using their respective image processors. We set a maximum generation length of 65,536 tokens across all models to accommodate the extended reasoning outputs of thinking models. To standardize response formatting, we append the instruction ``Give the final answer in \texttt{<answer>} \texttt{</answer>} tags.'' to every question. Models have no access to external retrieval, tools, or nautical-chart references during inference.
\section{Subtask Performance}
We provide a finer-grained breakdown of model performance within each \benchname capability level.

\subsection{ENC Perception}
We additionally evaluate Isolated and In-Context Symbol Recognition at the broad feature-category level. Unlike the original evaluation, which requires precise identification of nautical features and their characteristics, broad-category scoring accepts responses that correctly identify the general feature class without specifying its detailed attributes.

\begin{table*}[htbp]
    \centering \small
    \begin{tabular}{lcccccc}
    \toprule
    & \multicolumn{3}{c}{\textbf{Isolated Symbol Recognition}}
    & \multicolumn{3}{c}{\textbf{In-Context Symbol Recognition}} \\
    \cmidrule(lr){2-4}
    \cmidrule(lr){5-7}
    \textbf{Model}
    & \textbf{Full} & \textbf{Broad} & \textbf{Gain (pp)}
    & \textbf{Full} & \textbf{Broad} & \textbf{Gain (pp)} \\
    \midrule

    Qwen3.8-27B
    & 0.2 & 9.6 & +9.4
    & 7.3 & 23.0 & +15.7 \\

    LLaVA-OneVision2-8B
    & 0.0 & 15.4 & +15.4
    & 3.4 & 14.2 & +10.7 \\

    Qwen3.6-35B-A3B
    & 0.6 & 8.2 & +7.6
    & 7.2 & 25.1 & +17.8 \\

    Ovis2.6-30B-A3B
    & 0.4 & 12.3 & +11.9
    & 4.8 & 21.3 & +16.5 \\

    Qwen3VL-30B-A3B-Instruct
    & 0.0 & 6.2 & +6.2
    & 5.8 & 19.0 & +13.3 \\

    InternVL3.5-30B-A3B
    & 0.0 & 8.6 & +8.6
    & 3.7 & 14.1 & +10.4 \\

    KimiVL-16B-A3B-Thinking
    & 0.0 & 8.8 & +8.8
    & 3.1 & 11.4 & +8.3 \\

    KimiVL-16B-A3B-Instruct
    & 0.0 & 14.6 & +14.6
    & 2.6 & 13.2 & +10.5 \\

    Idefics3-8B-LLaMA3
    & 0.0 & 7.4 & +7.4
    & 2.7 & 9.6 & +6.9 \\

    LLaVA-OneVision-7B
    & 0.0 & 15.4 & +15.4
    & 3.0 & 15.5 & +12.5 \\

    \bottomrule
    \end{tabular}
    \caption{Comparison of full and broad-category recognition accuracy (\%) for Isolated and In-Context Symbol Recognition. Full requires precise nautical feature identification, whereas Broad accepts correct general feature categories without requiring fine-grained characteristics. Gain (pp) denotes the percentage-point improvement from full to broad-category scoring.}
    \label{tab:broad-category-score}
\end{table*}

Table~\ref{tab:broad-category-score} compares accuracy under the original and broad-category scoring. 

\subsection{Spatial Reasoning}

\begin{table*}[htbp]
\centering \small
\setlength{\tabcolsep}{3.5pt}
\resizebox{\textwidth}{!}{%
\begin{tabular}{lccccccccccc}
\toprule
& \multicolumn{5}{c}{\textbf{Topological}}
& \multicolumn{3}{c}{\textbf{Metric}}
& \multicolumn{3}{c}{\textbf{Directional}} \\
\cmidrule(lr){2-6}
\cmidrule(lr){7-9}
\cmidrule(lr){10-12}

\textbf{Model}
& \textbf{Touch} & \textbf{Within} & \textbf{Cross}
& \textbf{Overlap} & \textbf{Equal}
& \textbf{Qualitative} & \textbf{Quantitative}
& \textbf{MAPE (\%)}$\downarrow$
& \textbf{Cardinal} & \textbf{Bearing}
& \textbf{Mean Dev. ($^\circ$)}$\downarrow$ \\
\midrule

Qwen3.8-27B
& 43.4 & \textbf{72.0} & \textbf{46.8} & \textbf{49.7} & 28.2
& \textbf{96.7} & \textbf{43.6} & \textbf{18.1}
& \textbf{74.0} & \textbf{92.2} & \textbf{5.1} \\

LLaVA-OneVision2-8B
& 35.0 & 41.4 & 28.1 & 30.2 & 33.2
& 86.4 & 2.7 & 68.6
& 47.0 & 13.6 & 60.1 \\

Qwen3.6-35B-A3B
& 41.6 & 54.8 & 46.5 & 37.5 & 28.0
& 92.8 & 15.0 & 34.9
& 52.9 & 39.9 & 12.3 \\

Ovis2.6-30B-A3B
& 38.6 & 48.2 & 39.4 & 44.3 & 30.8
& 78.7 & 2.5 & 53.0
& 49.0 & 26.1 & 25.1 \\

Qwen3VL-30B-A3B-Instruct
& 35.0 & 52.7 & 36.5 & 37.3 & 27.6
& 83.5 & 1.1 & 62.6
& 55.4 & 21.6 & 30.1 \\

InternVL3.5-30B-A3B
& \textbf{46.3} & 52.9 & 38.8 & 40.3 & 35.0
& 63.3 & 1.1 & 69.2
& 51.0 & 12.1 & 57.1 \\

KimiVL-16B-A3B-Thinking
& 44.6 & 55.0 & 44.4 & 32.2 & 28.6
& 72.8 & 1.5 & 64.5
& 55.2 & 17.5 & 51.4 \\

KimiVL-16B-A3B-Instruct
& 28.2 & 49.1 & 30.8 & 21.3 & 23.2
& 77.7 & 2.6 & 73.3
& 22.9 & 11.7 & 68.0 \\

Idefics3-8B-LLaMA3
& 29.1 & 47.1 & 41.5 & 29.0 & 39.1
& 61.5 & 1.4 & 85.2
& 22.2 & 8.0 & 77.0 \\

LLaVA-OneVision-7B
& 37.0 & 50.2 & 34.5 & 38.7 & \textbf{44.2}
& 71.4 & 0.3 & 84.2
& 26.4 & 5.1 & 98.0 \\

\bottomrule
\end{tabular}%
}
\caption{Fine-grained performance on Spatial Reasoning. Results are accuracy (\%), with \textbf{bold} marking the best score in each column. Quantitative metric counts a distance as correct within 10\% of the reference. Bearing counts a bearing as correct within $\pm10^\circ$. MAPE is the mean absolute percentage error of the stated distances, with each error capped at 100\%, and Mean dev.\ is the mean absolute angular error of the stated bearings (lower is better).}
\label{tab:supp-spatial}
\end{table*}

Spatial Reasoning performance also varies by relation type. In general, models achieve higher accuracy for qualitative distance and categorical direction than for exact numerical questions such as quantitative distance and exact bearings.

\section{LLM-as-Judge} \label{app:llm-as-judge}
We use GPT-5.6 Sol to evaluate model responses that do not match their reference answers after normalization. The judge assigns a binary correctness score based on semantic equivalence and answer specificity, accepting valid paraphrases while rejecting responses that omit essential distinctions or contradict the reference. The full evaluation prompt appears in Figure~\ref{fig:llm-judge-prompt}.

\clearpage
\onecolumn

\begin{figure}[!htbp]
\centering
\begin{judgeprompt}
You are judging whether a model's answer to a question about a nautical chart is correct.

Question: {question}
Reference answer: {reference_answer}
Model answer: {model_answer}

The reference answer is correct. Compare the model's final answer with it. Score 1 if they give the same answer, and 0 otherwise.

Rules:

Find the final answer the model commits to. It does not need to be inside answer tags. If the response ends without stating a final answer, score 0.

The wording may differ. Accept synonyms, paraphrases, equivalent units, and extra words, as long as the answer refers to the same thing as the reference.

Names and chart text must be spelled as in the reference. Ignore capitalization, punctuation, and spacing. A misspelled name is wrong. So is a name with words added or removed so that it refers to a different place or feature.

The answer is wrong if it is less specific than the reference, contradicts it, or gives more than one answer without choosing one.

Do not solve the question yourself.

Return only this JSON: {"score": 0 or 1, "reason": "<one sentence>"}
\end{judgeprompt}
\caption{Complete prompt used for LLM-as-Judge evaluation. The judge compares model responses with reference answers and returns a binary correctness score with a brief explanation.}
\label{fig:llm-judge-prompt}
\end{figure}
\twocolumn

\subsection{LLM-as-Judge Disagreement Analysis} \label{app:llm-as-judge-disagreement}
We examine disagreements between human annotations and LLM-as-Judge evaluations. Among the 93 responses with human consensus, GPT-5.6 Sol disagrees with the annotators on six cases.

\paragraph{Judge-Consensus Disagreements.}
For the reference answer \textit{Buckeye Reef} and model response \textit{Buckeye Reef 2}, both human annotators accept the response because it identifies the correct general location despite the additional `2,' while the LLM judge rejects it due to the mismatch in the feature name. Conversely, for the reference answer \textit{Big Island} and model response \textit{Big Island Passage}, the LLM judge accepts the response despite both human annotators marking it incorrect.

\paragraph{Human Annotation Disagreement.}
Disagreement also occurs between human annotators. For example, one annotator accepts \textit{contour line} as equivalent to \textit{depth contour}, while the other rejects it because \textit{contour line} does not specify that the contour represents depth. The LLM judge accepts the response.

\end{document}